\documentclass[11pt,a4paper]{article}
\usepackage[hyperref]{acl2019}
\usepackage{times}

\usepackage{latexsym}
\usepackage[plain]{fancyref}    
\usepackage{url}
\usepackage{amsmath}
\usepackage{enumitem}
\setlist[enumerate]{itemsep=0mm}

\newcommand{\astfootnote}[1]{
\let\oldthefootnote=\thefootnote

\setcounter{footnote}{0}
\renewcommand{\thefootnote}{\fnsymbol{footnote}}
\footnotetext{#1}
\let\thefootnote=\oldthefootnote
}
\aclfinalcopy 
\def\aclpaperid{127} 

\title{Zero-shot transfer for implicit discourse relation classification}

\author{
  Murathan Kurfal{\i} \\
  Department of Linguistics \\
  Stockholm University  \\
  \texttt{murathan.kurfali@ling.su.se} \\
  \\\And
  Robert {\"O}stling \\
  Department of Linguistics \\
  Stockholm University \\
  \texttt{robert@ling.su.se}
  }

\date{}

\begin{document}
\maketitle

\begin{abstract}
Automatically classifying the relation between sentences in a discourse is a challenging task, in particular when there is no overt expression of the relation. It becomes even more challenging by the fact that annotated training data exists only for a small number of languages, such as English and Chinese. We present a new system using zero-shot transfer learning for implicit discourse relation classification, where the only resource used for the target language is unannotated parallel text. This system is evaluated on the discourse-annotated TED-MDB parallel corpus, where it obtains good results for all seven languages using only English training data.
\end{abstract}

\section{Introduction}

The difference between a set of randomly selected sentences and a discourse lies in coherence. Among other attempts at defining the elusive nature of coherence, one way is to look at the meaning conveyed between the adjacent pair of sentences. In the current study, we follow the Penn Discourse Treebank (PDTB) framework which regards abstract objects \citep{asher2012reference} as the units of discourse and views the text as a collection of discourse level predicates, each taking two arguments. Such predicates, called discourse connectives, may (Ex.\ \ref{ex:exp}) or may not (Ex.\ \ref{ex:imp}) be represented in the surface form:
\begin{enumerate}
    \item \label{ex:exp} \underline{Because} \textbf{the drought reduced U.S. stockpiles}, \textit{they have more than enough storage space for their new crop}, and that permits them to wait for prices to rise.
    \item \label{ex:imp} \textit{But a few funds have taken other defensive steps. Some have raised their cash positions to record levels.} \underline{Implicit = BECAUSE} \textbf{High cash positions help buffer a fund when the market falls}.
\end{enumerate}
where \textit{italics} represents the first and \textbf{boldface} the second argument to the \underline{underlined} discourse connective.  The discourse relations which lack an overt discourse connective (Ex. \ref{ex:imp}) are referred as \emph{implicit discourse relations} and are shown to be the most challenging part of the discourse parsing \citep[e.g.][]{pitler2009automatic}. 

In this paper, we perform implicit discourse relation classification using three recent data sets annotated according to the same guidelines: Penn Discourse Treebank (PDTB) 3.0, the Turkish Discourse Bank (TDB), and the multilingual TED-MDB. To the best of our knowledge, multilingual training and zero-shot transfer has not previously been investigated for this problem.
The results suggest that an implicit discourse relation classifier can transfer well across dissimilar languages, and that pooling training data from unrelated languages (English and Turkish) leads to significantly better performance for all languages.

\section{Related Work}

Implicit discourse relation recognition is often handled as a classification task, where earlier studies focused on using linguistically rich features \cite{pitler2009automatic,zhou2010predicting, park2012improving,rutherford2014discovering}.

Recently, neural network approaches have become popular.
\citet{ji2015one} use two RNNs on the syntactic trees of the arguments whereas \citet{zhang2015shallow} use a CNN to perform discourse parsing in a multi-task setting where they consider both explicit and implicit discourse relations. 

\citet{rutherford2016robust} use a simple yet robust feedforward network and achieves the highest performance on the out-of-domain blind test in the CoNLL 2016 shared task \citep{xue2016conll}. 

\citet{lan2017multi} apply a multi-task attention-based neural network model whereas \citet{bai2018deep} focus on the representation of the sentence pair and take different levels of text, from character to sentence pair, into account to achieve a richer representation.

\citet{dai2018improving} adopt a similiar approach and represent discourse units by considering a wider paragraph-level context. The discourse unit representations are created by a Bi-LSTM which takes a sequence of discourse relations in a paragraph which enables capturing the inter-dependencies between discourse relations as well. 

\section{Data}

We use four different data sets: the Penn Discourse Treebank (PDTB) version 2.0 \cite{prasad2008penn} and version 3.0 \cite{prasad2018discourse}, as well as the TED Multilingual Discourse Bank (TED-MDB, \citealt{zeyrekted}) and the Turkish Discourse Bank (TDB, \citealt{zeyrek2017tdb}).

The PDTB is built upon the 1 million word Wall Street Journal corpus and is the largest available resource for discourse relations. 
Most related work uses PDTB 2.0, so we include this for comparing our baseline to previous work.

The recently released PDTB 3.0 adopts a new annotation schema as well as an updated sense hierarchy. PDTB 3.0 includes the annotations of PDTB 2.0 updated according to the new annotation schema, as well as about 13 thousand new annotations, of which about 5K are implicit relations \citep{prasad2018discourse}. The distribution of the top level senses of the implicit discourse relations in both PDTB versions is provided in \Fref{tab:pdtb-stats}.

TED-MDB \citep{zeyrekted} is the first parallel corpus annotated for discourse relations. It closely follows the PDTB 3.0 framework and includes the manual annotations of six TED talks in seven languages (English, Turkish, European Portuguese, Polish, German, Russian) aiming to allow crosslingual comparison of discourse relations\footnote{The TED-MDB annotations are available at: https://github.com/MurathanKurfali/Ted-MDB-Annotations}. It has recently also been updated with Lithuanian  \citep{oleskevicieneobservations}.

Despite the high number of languages covered by TED-MDB, the amount of annotated text per language is limited (see Table \ref{tab:tdb-tedmdb-stats}). Therefore, in the current study, we limit ourselves with the top level senses, namely Expansion, Contingency, Comparison and Temporal. We only use TED-MDB for evaluation.

Among the TED-MDB languages other than English, only Turkish has another corpus annotated with PDTB 3.0 discourse annotations, namely the Turkish Discourse Bank (TDB). TDB is a multi-genre corpus of 40~000 words, considerably less than the PDTB (see Table \ref{tab:tdb-tedmdb-stats}), but it provides the only directly comparable baseline to assess the performance of zero-shot learning.

\begin{table}[h]
\centering
\small
\begin{tabular}{lrrr|rrr}
\hline
&\multicolumn{3}{c|}{PDTB2}&\multicolumn{3}{c}{PDTB3}\\ \hline
Sense& Train & Dev  & Test & Train & Dev  & Test \\ \hline
Comp. & 1894 & 401  &146   & 1828&404 & 153  \\
Cont. &  3281& 628 & 276 &5872 &1159 & 527  \\
Exp. & 6792 & 1253 & 556  & 7939 &1466  & 643 \\
Temp. & 665 & 93 & 68 & 1413 &230 & 148  \\ \hline
\end{tabular}
\caption{Distribution of top level senses of the implicit discourse relations in PDTB 2.0 and PDTB 3.0 training, development and test sets: comp(arison), cont(ingency), exp(ansion), temp(oral).}
\label{tab:pdtb-stats}
\end{table}

\begin{table*}[h]
\centering
\begin{tabular}{lccccc}
\hline
Language & Comparison & Contingency & Expansion & Temporal & Total \\ \hline
English &20 (10.31\%)&52 (26.80\%)&107 (55.15\%)&15 (7.73\%) & 194 (100\%)\\
German &13 (6.07\%) &41 (19.16\%)&148 (69.16\%)&12 (5.61\%) & 214 (100\%)\\
Lithuanian &26 (10.57\%)&53 (21.54\%)&154 (62.60\%)&13 (5.28\%) & 246 (100\%)\\
Polish &19 (9.74\%)&28 (14.36\%)&130 (66.67\%)&18 (9.23\%) & 195 (100\%)\\
Portuguese &23 (9.06\%)&47 (18.50\%)&169 (66.54\%)&15 (5.91\%) & 254 (100\%)\\
Russian &16 (7.24\%)&31 (14.03\%)&169 (76.47\%)&5 (2.26\%) & 221 (100\%)\\
Turkish &20 (9.90\%)&29 (14.36\%)&140 (69.31\%)&13 (6.44\%) & 202 (100\%)\\ \hline
TDB (training) &71 (10.94\%)&142 (21.88\%)&363 (55.93\%)&73 (11.25\%) & 649 (100\%)\\
TDB (dev) &11 (9.82\%)&31 (27.68\%)&49 (43.75\%)&21 (18.75\%) & 112 (100\%)\\  \hline
\end{tabular}
\caption{Distribution of top level senses of the implicit discourse relations in the TED-MDB and TDB corpora. The numbers within the parenthesis indicate the ratio. Since there is no official training/dev split for TDB, we arbitrarily chose two sections with different genres for the development set.}
\label{tab:tdb-tedmdb-stats}
\end{table*}

\section{Model}

The main purpose of this study is to assess the performance of transfer learning on the implicit discourse relation classification task. To this end, we use a simple feedforward network fed with multilingual sentence embeddings following the finding of \cite{rutherford2017systematic} which shows that simple discourse models with feedforward layers perform on par or better than those of with surface features or recurrent and convolutional architectures.

We follow the model of \cite{rutherford2016robust} due to its simplicity and robust nature even in the multilingual setting with different argument and discourse relation representations. We represent the arguments of the discourse relation via pre-trained LASER model \cite{artetxe2018massively}. 
LASER is chosen as it is the current state-of-the-art model on several Natural Language Inference (NLI) transfer learning tasks, a sentence relation classification problem similar to discourse relation classification.

Given the argument vectors, $V_{arg1}$ and $V_{arg2}$, the next step is to represent the discourse relation in a way that the interactions between them are captured. To this end, we model the discourse relation vector, $V_{dr}$, by performing the following  pair-wise vector operations following the DisSent model of \cite{nie2017dissent}:

    $$ V_{avg} = \dfrac{1}{2} ( V_{arg1} + V_{arg2} ) $$
    $$ V_{sub} =  V_{arg1} - V_{arg2} $$
    $$ V_{mul} =  V_{arg1} * V_{arg2} $$
    $$ V_{dr} = [ V_{arg1}, V_{arg2},V_{avg},V_{sub}, V_{mul} ]   $$

The resulting vector is further fed into a hidden layer $h_t$ with $d$ hidden units\footnote{We use d=100 in the experiments } to achieve a more abstract representation of the relation and finally the output $o$ is calculated using the sigmoid function. This model is also essentially the same as was used by \citet{artetxe2018massively} for NLI transfer learning.

\section{Experiments}

We formulate the implicit relation classification as four "one vs other" binary classification task. We follow the conventional setting of the first study \cite{pitler2009automatic} and split the PDTB 2.0 into training (sections 2-20), development (sections 0-1 and 23-24) and test sets (sections 21-22) to have directly comparable results with the previous work. However, following the PDTB's original distinction but unlike some previous work, we distinguish Entity-based relations from implicit relations. Each classifier is trained on an equal number of positive and negative instances where the negative instances are randomly selected in each epoch to have a better representation of the data during the training. This model is evaluated on the PDTB 2.0 test set to confirm whether our model performs adequatly on same-language, same-domain data. These results are directly comparable to previous work.

As TED-MDB is annotated according to the PDTB 3.0 framework, we train separate classifiers on PDTB 3.0 following the same convention as above. We test the trained models on all the implicit discourse relations in the TED-MDB corpus. 

The PDTB framework  allows annotations to be labelled with more than one label. In such cases we only keep the first label, in line with previous studies \citep[among others][]{ji2015one, rutherford2017systematic}.

The argument vectors are kept fixed during the training, and we do not update the parameters of the LASER model. We use cross-entropy loss, and AdaGrad as the optimizer. We evaluate using the model which achieved the highest F-score on the development set. As for the regularization, we use a dropout layer between the input and the hidden layer with a dropout probability of 0.3. All models are run 100 times to estimate the variance due to random initialization and stochastic training. All the models are implemented in PyTorch\footnote{https://pytorch.org/}. 
\begin{table*}[h]
\centering
\begin{tabular}{lcccc}
\hline
 & Comparison & Contingency & Expansion & Temporal \\
\cite{pitler2009automatic} &21.96& 47.13& -& 16.76\\
\cite{zhou2010predicting} &31.79& 47.16& 70.11& 20.30\\
\cite{park2012improving}  &31.32 &49.82 & - & 26.57 \\
\cite{rutherford2014discovering}&39.70& 54.42 &70.23&28.69\\
\cite{zhang2015shallow}&33.22& 52.04&69.59&30.54\\
\cite{ji2015one}&35.93&52.78&-&27.63\\
\cite{lan2017multi}& 40.73 &\textbf{58.96} &\textbf{72.47} &38.50\\
\cite{bai2018deep}& \textbf{47.85 }& 54.47&  70.60 & 36.87\\
 \cite{dai2018improving}&46.79& 57.09& 70.41& \textbf{45.61}\\  \hline
Baseline&24.49&41.75&69.41&12.20 \\ 
 \hline
Our system& 28.19 ($\pm$0.83) & 50.63 ($\pm$1.00) & 64.07 ($\pm$1.90) & 29.22 ($\pm$2.53)\\
\hline
\end{tabular}
\caption{Comparison of the F scores (\%) of binary classifiers on PDTB 2.0 test set. Left out scores refer to the results where EntRel relations are also considered to be Expansion.}
\label{tab:pdtb2}
\end{table*}

\section{Results and Discussion}
\begin{table*}[]
\centering
\begin{tabular}{lccccc}
\hline
Language&Comparison&Contingency&Expansion&Temporal&Average\\ \hline
Baseline (PDTB 3.0)  &18.84&52.75&60.83&18.28 & 37.67  \\ 
\hline
PDTB 3.0 & 24.90 ($\pm$0.87) & 59.18 ($\pm$0.72) & 60.10 ($\pm$1.32) & 36.73 ($\pm$1.45) &45.23 \\
\hline
German& 8.62 ($\pm$1.61) & 37.34 ($\pm$1.43) & 70.81 ($\pm$3.16) & 40.11 ($\pm$4.32) &39.22 \\
English& 10.18 ($\pm$3.31) & 40.92 ($\pm$1.80) & 62.28 ($\pm$2.16) & 50.45 ($\pm$5.26) &40.96 \\
Lithuanian& 23.50 ($\pm$2.33) & 34.64 ($\pm$1.43) & 62.35 ($\pm$2.65) & 36.78 ($\pm$3.28) &39.32 \\
Polish& 16.50 ($\pm$3.51) & 29.19 ($\pm$1.36) & 60.32 ($\pm$2.84) & 44.17 ($\pm$3.37) &37.54 \\
Portuguese& 19.59 ($\pm$1.99) & 33.85 ($\pm$1.27) & 66.83 ($\pm$2.57) & 37.04 ($\pm$3.43) &39.33 \\
Russian& 14.90 ($\pm$2.07) & 26.76 ($\pm$1.08) & 70.06 ($\pm$3.97) & 28.28 ($\pm$4.41) &35.00 \\
Turkish& 10.99 ($\pm$3.16) & 25.28 ($\pm$1.23) & 64.14 ($\pm$2.96) & 33.66 ($\pm$4.31) &33.52 \\

\hline
\end{tabular}
\caption{F scores (\%) when the model is trained only on PDTB 3.0. In the table, PDTB 3.0 refers to the test set of the PDTB 3.0 corpus. The remaining rows refer to evaluations using TED-MDB. }
\label{tab:pdtb3_res}

\end{table*}

\Fref{tab:pdtb2} shows the same-language, same-domain performance of our system, in comparison to previous work. All figures refer to PDTB 2.0 test set F-score, when trained on the PDBT 2.0 training set, and are directly comparable. While our model does not achieve state-of-the-art performance in this setting, this experiment shows that it performs adequately for English, and provides a reasonable baseline for the zero-shot experiments presented in Tables \ref{tab:pdtb3_res} and \ref{tab:tdb-effect}. We also include a naive baseline system which always predicts \textsc{true} and is evaluated on the respective (PDTB 2.0 or PDTB 3.0) test set in our comparisons.

In all zero-shot experiments, evaluation is performed on the available test data with PDTB 3.0 annotations: TED-MDB, and the PDTB 3.0 test set itself. Results in \Fref{tab:pdtb3_res} use PDTB 3.0 only for training, whereas \Fref{tab:tdb-effect} presents the effect of having additional training data from Turkish (a language unrelated to English). Pooling training data from different languages is possible since our model is language-agnostic.


In all zero-shot experiments, we see similar levels of performance across all the evaluated languages in TED-MDB. While not completely comparable numerically since annotations differ slightly between languages, this evaluation set consists of parallel sentences annotated according to the same guidelines. The similarity in scores between the training language(s)---English and/or Turkish---and the remaining languages indicates that little accuracy is lost during transfer.

Comparing the performances with and without additional Turkish data, TDB, reveals that adding a small amount (relative to the size of PDTB 3.0) of Turkish training data improves the F-scores by a statistically significant amount\footnote{On a sense-wise analysis, we observe that the main increase is in the Expansion relations; however, there is no decrease in any of the other senses. } for not only Turkish, but for all the languages in TED-MDB \Fref{tab:tdb-effect}.



\section{Conclusion}

In the current paper we have presented the (to the best of our knowledge) first study of zero-shot learning in the implicit discourse relation classification task. Our method does not require any discourse level annotation for the target languages, yet still achieves good performance even for those languages where no training data is available. The performance is further increased by pooling training data from multiple languages. Using our published code\footnote{https://github.com/MurathanKurfali/multilingual\_IDRC} and publicly available resources it can used for implicit discourse classification in nearly a hundred languages.

\begin{table}[]
\centering
\begin{tabular}{lccc}
\hline
Language&TDB&PDTB3&PTDB3+TDB\\ 
\hline
PDTB3 Test&35.35&45.23&\textbf{45.62} \\ \hline
German&36.93&39.22&\textbf{41.44} \\
English&38.06&40.96&\textbf{42.22} \\
Lithuanian&36.92&39.32&\textbf{41.94} \\
Polish&35.48&37.54&\textbf{39.65} \\
Portuguese&37.58&39.33&\textbf{41.04} \\
Russian&30.92&35.00&\textbf{38.23} \\
Turkish&\textbf{39.58}&33.52&37.14 \\

\hline
\end{tabular}
\caption{Comparison of average F-scores (\%) when the model is trained on different training sets. Bold means significantly higher F-score than the second highest column ($p < 0.001$, Mann-Whitney U test).}
\label{tab:tdb-effect}

\end{table}

\section*{Acknowledgments}
We would like to thank Bonnie Webber for her help in obtaining PDTB 3.0 and Mats Wir\'en for his useful comments. 

\bibliography{acl2019}
\bibliographystyle{acl_natbib}

\end{document}